\documentclass[sigconf]{acmart}
\AtBeginDocument{%
  }

\copyrightyear{2026}
\acmYear{2026}
\setcopyright{cc}
\setcctype{by}
\acmConference[MM '26]{Proceedings of the 34th ACM International Conference on Multimedia}{November 10--14, 2026}{Rio de Janeiro, Brazil}
\acmBooktitle{Proceedings of the 34th ACM International Conference on Multimedia (MM '26), November 10--14, 2026, Rio de Janeiro, Brazil}
\acmISBN{979-8-4007-2213-4/2026/11}
\acmDOI{10.1145/3767308.3835308}
\usepackage{multirow}
\usepackage{colortbl}
\definecolor{rowgray}{gray}{0.9}
\begin{document}

\title{UniAR: A Unified Framework for Autism Recognition Enhanced by Multi-View Prompt Learning}


\author{Lei Xin}
\authornote{The first three authors contributed equally to this work.}
\affiliation{
  \institution{Wuhan University}
  \city{Wuhan}
  \state{Hubei}
  \country{China}}
\email{2835838600@qq.com}

\author{Zeheng Wang}
\authornotemark[1]
\orcid{0009-0009-2658-7587}
\affiliation{
  \institution{Northeast Normal University}
  \city{Changchun}
  \state{Jilin}
  \country{China}}
\email{wangzeheng624@nenu.edu.cn}

\author{Jiayin Zhu}
\authornotemark[1]
\orcid{0000-0001-7226-4791}
\affiliation{
\institution{The Hong Kong University of Science and Technology (Guangzhou)}
\city{Guangzhou}
\state{Guangdong}
\country{China}}
\email{jzhu709@connect.hkust-gz.edu.cn}

\author{Shihong Huang}
\orcid{0009-0002-0803-3846}
\affiliation{
  \institution{Nanjing Agricultural University}
  \city{Nanjing}
  \state{Jiangsu}
  \country{China}}
\email{23123122@stu.njau.edu.cn}

\author{Fanhu Zeng}
\affiliation{
  \institution{Imperial College London}
  \city{London}
  \country{United Kingdom}}
\email{challengezengfh@gmail.com}

\author{Changjiang Jiang}
\affiliation{
  \institution{Wuhan University}
  \city{Wuhan}
  \state{Hubei}
  \country{China}}
\email{jiangcj@whu.edu.cn}

\author{Dengbo He}
\orcid{0000-0003-4359-4083}
\affiliation{
  \institution{The Hong Kong University of Science and Technology}
  \city{Hong Kong}
  \country{China}}
\email{dengbohe@ust.hk}

\author{Yutao Yue}
\orcid{0000-0003-4532-0924}
\affiliation{
  \institution{The Hong Kong University of Science and Technology (Guangzhou)}
  \city{Guangzhou}
  \state{Guangdong}
  \country{China}}
\email{yutaoyue@hkust-gz.edu.cn}

\author{Zhenglun Kong}
\authornote{Zhenglun Kong is the corresponding author.}
\affiliation{
  \institution{Harvard University}
  \city{Boston}
  \country{USA}}
\email{zhenglun\_kong@hms.harvard.edu}

\renewcommand{\shortauthors}{Lei Xin et al.}

\begin{abstract} Autism Spectrum Disorder (ASD) is a complex neurodevelopmental disorder for which early and accurate diagnosis is critical to improving long-term developmental outcomes. However, existing ASD recognition methods are often constrained by the scarcity of diagnostic text data, forcing them to rely mainly on visual analysis and limiting their ability to model clinically meaningful semantic reasoning. To address this challenge, we propose UniAR, a unified framework enhanced by multi-granularity prompt learning for robust ASD recognition under heterogeneous data variations. Specifically, UniAR leverages a large multimodal model to generate hierarchical diagnostic descriptions at the word, phrase, and sentence levels, compensating for the lack of paired clinical reports. To align the generated semantics with visual evidence, we further design a Mixture-of-Experts-based Multi-Scale Alignment Module, which dynamically matches vector-quantized visual prototypes with semantic representations at corresponding granularities. Extensive experiments on four benchmarks covering brain MRI and facial expression scenarios show that UniAR consistently outperforms existing state-of-the-art methods, achieving average accuracies of 75.9\% on MRI benchmarks and 91.6\% on facial benchmarks, while improving average Accuracy on MRI benchmarks by 1.5 percentage points and average Accuracy on facial benchmarks by 1.2 percentage points over baselines. 
These results demonstrate that UniAR offers a robust and interpretable framework for ASD screening under semantic scarcity.

\end{abstract}



\maketitle
\section{Introduction}

Autism Spectrum Disorder (ASD) represents a complex neuro-developmental condition characterized by core deficits in social communication and interaction, alongside restricted, repetitive patterns of behavior. According to statistics from the World Health Organization (WHO), the global average prevalence of ASD in children is approximately 1\%~\cite{salari2022global}. A substantial body of research indicates that early childhood, typically defined as 0–6 years, serves as a critical window for neurodevelopment; achieving early identification and intervention during this phase is instrumental in improving long-term developmental outcomes~\cite{sandbank2023autism,dawson2012early,santomauro2025global}. Consequently, enhancing the accessibility and accuracy of early ASD diagnosis holds significant social and public health implications~\cite{santomauro2025global}.

In clinical practice, screening for autism spectrum disorder (ASD) relies on standardized scales (\textit{e.g.}, M-CHAT-R/F) and parent reports~\cite{robins2014validation}. However, these subjective measurement methods often struggle to address the inherent heterogeneity of ASD presentation~\cite{wieckowski2023sensitivity,robins2014validation,hobson2021moving,lord2019recognising}. Although machine learning has been applied to automatically identify ASD through behavioral imaging, existing methods primarily focus on unimodal visual analysis. Therefore, they overlook crucial semantic descriptors in clinical diagnosis, such as emotional responses and social interactions~\cite{uddin2024deep}.

\begin{figure}[t]
    \centering
    \includegraphics[width=0.9\linewidth]{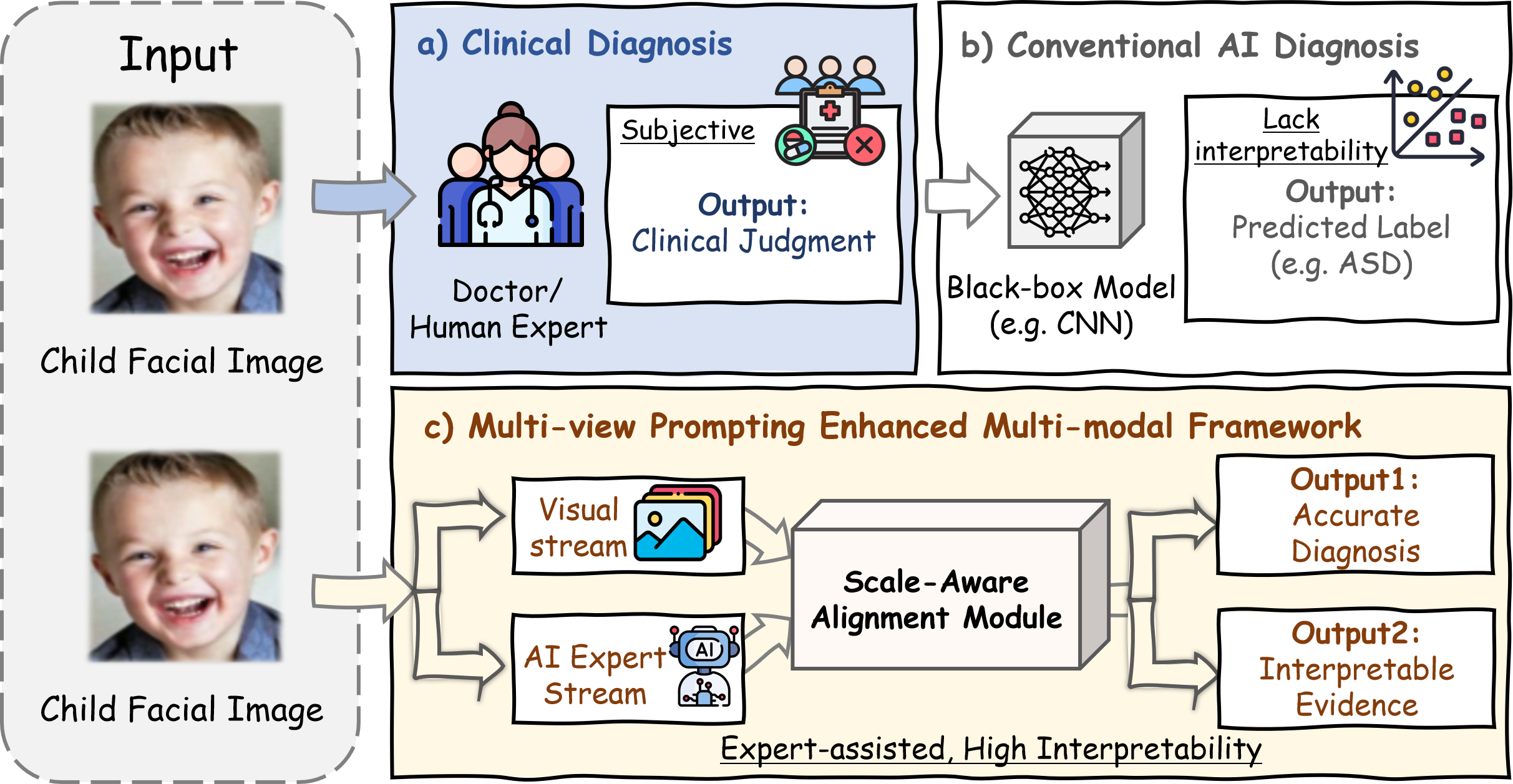}
    \caption{The framework comparisons.Existing autism recognition methods (a-b) are mainly implemented by domain experts or neural networks. In contrast, UniAR (c) integrates expert knowledge with neural networks to provide a more interpretable recognition solution, which is capable of outputting both diagnostic reports and results simultaneously.}
    \Description{Comparison of clinical diagnosis, conventional black-box AI diagnosis, and the proposed multi-view prompting framework, which produces both an ASD prediction and interpretable evidence.}
    \label{fig:first}
\end{figure}

While multimodal learning~\cite{khattak2023maple,zeng2025modalprompt} offers a promising avenue for integrating visual and semantic data, its application faces several obstacles. First, high-quality paired visual-text data is scarce and expensive to obtain~\cite{wu2024deep}. Furthermore, strict privacy regulations surrounding sensitive diagnostic text severely limit data sharing~\cite{michelson2022navigating,ganggayah2025accelerating}. These limitations prevent current multimodal methods from fully realizing their potential, thus hindering the deployment of accurate and practical diagnostic systems.

To address the challenges of data limitations and missing text modalities, we propose \textbf{UniAR, a Multi-Granularity Prompt Learning framework based on Cross-modal Alignment}, as shown in Figure~\ref{fig:first}. Specifically, we utilize generative semantic augmentation to automatically synthesize multi-granularity descriptive text from visual inputs, effectively compensating for the scarcity of real diagnostic reports. Recognizing the complexity of ASD symptoms, we design a hierarchical ``word-phrase-sentence'' alignment mechanism that maps semantic descriptions to pixel-level, regional, and global visual features, respectively. By linking diagnostic semantics with behavioral biomarkers at different granularities, UniAR enables effective multi-scale visual-semantic fusion.

We conduct extensive experimental validation of UniAR on four heterogeneous public benchmarks, covering brain MRI connectivity (ABIDE I \& II) and facial behavioral expressions (HRM \& Kaggle). While these public benchmarks mainly provide standard binary ASD-vs-control evaluation settings, they are limited in reflecting fine-grained symptom severity and real-world distribution shifts encountered in practical screening. \textbf{To address this limitation, we further construct the first cross-platform social-media autism benchmark, termed ASD-MM,} which consists of autism-related videos paired with expert-associated diagnostic descriptions and formulates autism-related assessment as a four-class problem (\textit{i.e.}, Normal, Mild, Moderate, and Severe). Experimental results show that UniAR consistently outperforms strong single-modal and multimodal baselines, achieving average accuracies of 75.9\% on MRI benchmarks and 91.6\% on facial benchmarks. UniAR also achieves the best performance on the challenging ABIDE I and dynamic HRM datasets. Moreover, on the self-constructed ASD-MM benchmark, UniAR consistently improves over the EAC baseline across all evaluated social platforms, further demonstrating its robustness and practical generalization.


In summary, our paper makes the following contributions:
\begin{itemize}
    \item We propose UniAR, a unified visual-semantic alignment framework for robust ASD recognition that transforms low-level visual biomarkers into interpretable high-level diagnostic insights through multi-granularity prompting and semantic-guided modeling.
    \item We develop a hierarchical word-phrase-sentence visual-semantic alignment mechanism that associates visual biomarkers with diagnostic semantics at multiple granularities, and further enhance visual representation learning through codebook-based prototype refinement, thereby improving robustness and diagnostic reasoning under limited data.
    \item Extensive experiments on public MRI and facial expression benchmarks, as well as a cross-platform social-media evaluation setting, demonstrate that UniAR consistently outperforms strong baselines while showing promising robustness and semantic interpretability.
\end{itemize}

\section{Related Work}
\begin{figure*}[t]
  \centering
  \includegraphics[width=0.8\textwidth]{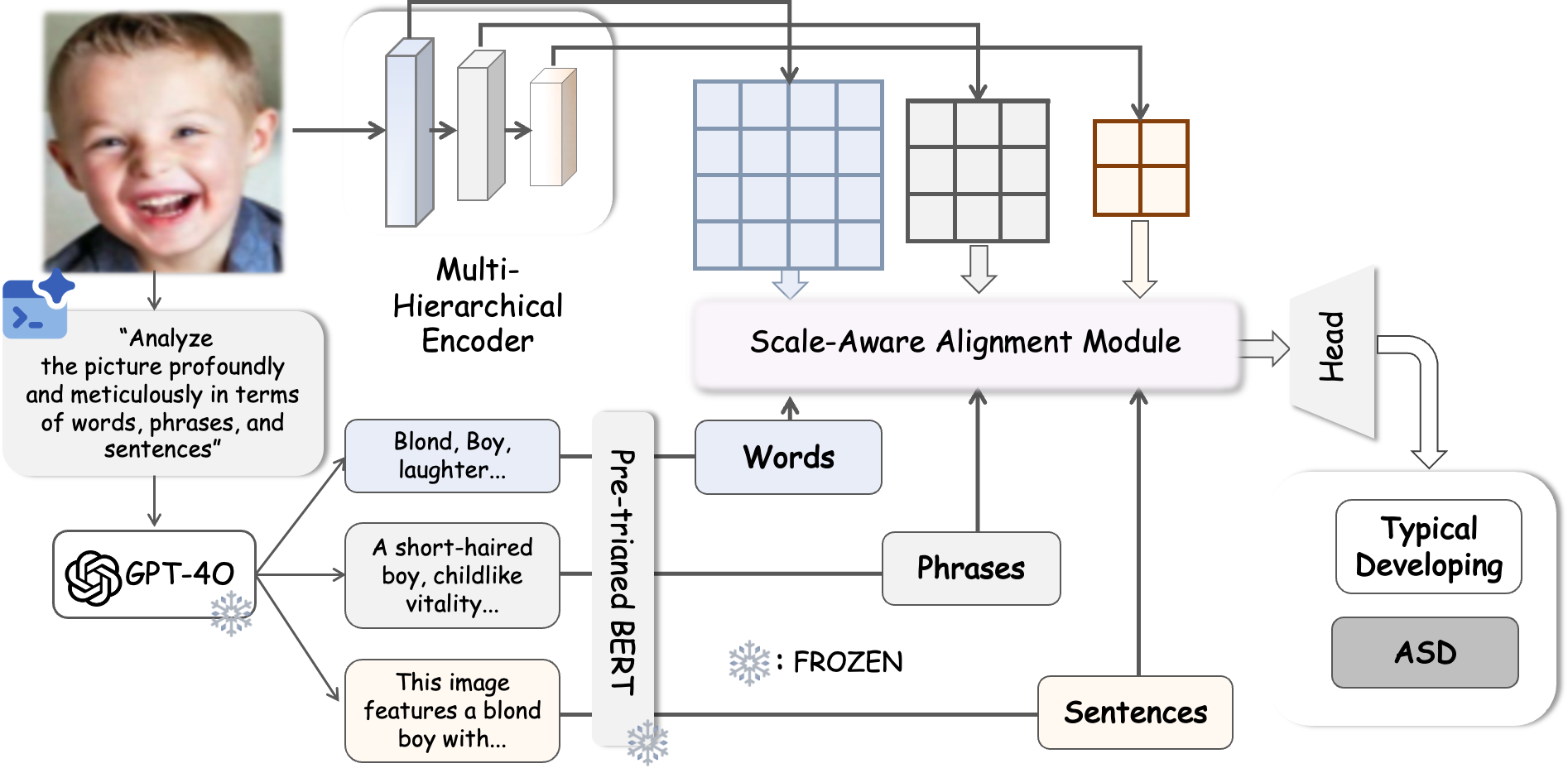} 
  \caption{ Overview of UniAR. The visual stream extracts multi-scale visual prototypes via a learnable visual codebook-augmented multi-hierarchical encoder. The semantic stream uses hierarchical prompts with GPT-4o to generate word-, phrase- and sentence-level descriptions, encoded by a frozen text encoder. A Scale-Aware Alignment Module aligns and fuses these multi-modal features for final classification. 
  }
  \Description{Pipeline of UniAR, including hierarchical visual encoding, AI-generated word-, phrase-, and sentence-level diagnostic descriptions, multi-scale cross-modal alignment, and final ASD classification.}
  \label{fig:framework}
\end{figure*}

\noindent\textbf{ASD Recognition.} Automated ASD recognition can be broadly grouped into neuroimaging-based diagnosis, facial-behavior analysis, and emerging multimodal systems. In neuroimaging, recent review papers indicate that MRI-based ASD diagnosis remains a major research paradigm, with most studies focusing on stronger discriminative modeling of resting-state or structural biomarkers on datasets such as ABIDE~\cite{uddin2024deep,schielen2024diagnosis,ganggayah2025accelerating}. Representative methods include connectome-based predictive modeling, graph-based modeling of functional connectivity, and more recent transformer or spectral architectures for ASD-related brain analysis~\cite{shen2017using,kim2021learning,li2021braingnn,bedel2023bolt,duan2024spectral,duan2025causal,liu2024made}. In parallel, facial and behavioral approaches identify ASD-related cues from head pose, facial expressions, and social-behavioral responses in images or videos~\cite{liu2024made,deng2024language,ismail2025multiscale}. Their implementations often build on robust FER techniques such as region attention, uncertainty modeling, and noisy-label learning~\cite{wang2020region,wang2020suppressing,she2021dive,zhang2021relative,zhang2022learn}. Some work further explores multimodal evidence such as audio-visual interaction modeling~\cite{zhao2025av}. Prior ASD-specific methods can be summarized as connectome modeling, behavioral cue modeling, and multimodal fusion, yet they mainly strengthen discriminative representations while explicit clinical-semantic modeling remains limited.

\noindent\textbf{Semantic Reconstruction and Missing-Modality Learning.} Another related line addresses limited supervision through augmentation, modality completion, or prompt-based adaptation. Classical augmentation improves generalization by enriching the sample space~\cite{shorten2019survey}, and recent neuroimaging work has extended this idea to fMRI-specific augmentation and computation toolchains~\cite{fang2025action}. In ASD settings, ASD-GANNet and GARL enhance robustness by synthesizing pseudo-samples or enriched feature distributions for small-data regimes~\cite{khan2024asd,zhou2024advancing}. More broadly, missing-modality learning studies how to preserve performance when one or more modalities are absent, typically through imputation, distillation, or robust fusion~\cite{wu2024deep,wang2023learnable}. Prompt-based multimodal adaptation~\cite{khattak2023maple,zeng2025local} further shows that learnable prompts can improve cross-modal conditioning and transfer~\cite{guo2025hide,zhao2025mllm}. DeFakerOne~\cite{DeFakerOne} introduces a industrial-grade image detection framework, demonstrating the feasibility leveraging massive datasets. Taken together, these methods can be summarized as sample-space augmentation, feature or modality completion, and prompt-conditioned adaptation. However, they mainly target robustness under incomplete inputs or limited labels, whereas diagnosis-level semantic reconstruction under missing clinical text remains underexplored. Fake-HR1~\cite{fakehr1} and DS-Net~\cite{qu2026detect} is based on question difficulty and thequality of dataset to achieve adaptive perception training.

\noindent\textbf{Medical Visual-Language Alignment.} Medical vision-language learning has progressed from coarse report-level matching to increasingly fine-grained semantic alignment. At a conceptual level, multi-grained vision-language pretraining showed the value of matching textual units with visual concepts across granularities~\cite{zeng2021multi}. In the medical domain, GLoRIA aligns report words with image subregions for label-efficient representation learning; MGCA models region-, instance-, and disease-level correspondences between medical images and reports; and MedKLIP introduces structured medical knowledge to align entity-level descriptions with image patches for diagnosis and grounding~\cite{huang2021gloria,wang2022multi,wu2023medklip}. More recent work extends this trend to brain-domain vision-language pretraining and aspect-level disease semantic decomposition~\cite{monajatipoor2024medical,phan2024decomposing}. Ivy-Fake~\cite{jiang2025ivy} and TwSG~\cite{jiang2026think} pioneered an explainable dataset for image understanding. Overall, existing methods can be summarized as report-level matching, multi-granularity local-global alignment, and knowledge-guided semantic grounding. However, they generally assume that paired reports or curated disease descriptions are available during training, leaving report-absent medical settings insufficiently studied.

\section{The Proposed Method}

\subsection{Preliminaries}
ASD recognition often suffers from \emph{semantic scarcity}: visual observations are available, while fine-grained clinical descriptions are usually missing. UniAR addresses this problem by modeling visual evidence and the latent diagnostic semantics inferred from it.

\noindent\textbf{Problem Formulation.}
We formulate ASD recognition as a supervised classification problem. Let $\mathcal{D}=\{(x_i,y_i)\}_{i=1}^{N}$ denote a dataset, where $x_i \in \mathcal{X}$ is the visual input and $y_i \in \mathcal{Y}$ is the diagnostic label. Since explicit clinical semantics are typically unavailable, we approximate them using a large multimodal generator $\mathcal{G}$ and optimize $\theta^*=\arg\max_{\theta}\sum_{i=1}^{N}\log P\big(y_i \mid x_i,\mathcal{G}(x_i);\theta\big)$, where $\theta$ denotes the learnable parameters of UniAR.

\noindent\textbf{Loss Function.}
The overall objective combines classification, semantic alignment, hierarchical regularization, fusion quality, and prototype learning: $\mathcal{L}_{\text{total}} = \mathcal{L}_{\text{cls}} + \alpha \mathcal{L}_{\text{align}} + \beta \mathcal{L}_{\text{sep}} + \gamma \mathcal{L}_{\text{fusion}} + \lambda_v \mathcal{L}_{\text{vq}}$,
where $\mathcal{L}_{\text{cls}}$ supervises ASD prediction, $\mathcal{L}_{\text{align}}$ enforces cross-modal consistency, $\mathcal{L}_{\text{sep}}$ preserves complementary hierarchical cues, $\mathcal{L}_{\text{fusion}}$ regularizes multi-scale fusion, and $\mathcal{L}_{\text{vq}}$ stabilizes prototype learning. 

Based on this formulation, UniAR is designed to reconstruct missing diagnostic semantics from visual evidence and align them with multi-scale visual representations for robust ASD recognition.



\subsection{Framework Overview}
\label{sec:framework}
ASD recognition under semantic scarcity involves three coupled challenges. First, raw visual observations are highly sensitive to subject-specific appearance variations, acquisition noise, and site-specific heterogeneity, making the learned representations unstable across samples. Second, although visual data may contain discriminative behavioral biomarkers, explicit clinical semantic supervision is typically unavailable, preventing the model from linking visual evidence to interpretable diagnostic concepts. Third, even when semantic descriptions are constructed, visual and textual representations remain mismatched in both feature distribution and abstraction level, making direct multimodal fusion suboptimal.

To address these issues, we propose UniAR, a unified visual-semantic framework for ASD recognition, as illustrated in Figure~\ref{fig:framework}. The framework follows a three-stage pipeline. First, a \textit{visual representation construction stage} transforms raw visual inputs into prototype-based multi-scale features through a hierarchical encoder and a learnable visual codebook, thereby reducing instance-specific noise while preserving ASD-relevant patterns. Second, a \textit{semantic construction stage} leverages a large multimodal model to generate hierarchical diagnostic descriptions at the word, phrase, and sentence levels, providing explicit semantic anchors for otherwise purely visual evidence. Third, a \textit{scale-aware alignment stage} progressively aligns visual prototypes with semantic embeddings at matched granularities, enabling robust cross-modal interaction and final ASD prediction.

In this way, UniAR does not treat multimodal learning as a simple combination of visual and textual features. Instead, it first stabilizes the visual representation, then reconstructs missing diagnostic semantics, and finally resolves the scale mismatch between the two modalities through explicit alignment. This design allows the framework to jointly improve classification robustness and diagnostic interpretability.

\subsection{Multi-Granularity Visual-Semantic Representation Construction Module}
\label{sec:representation}
Following the framework overview, we first construct the multi-granularity visual-semantic representation used by UniAR. Effective ASD recognition requires both robust visual representations and semantically grounded diagnostic cues. However, raw visual features are often entangled with subject-specific variations and irrelevant noise, while stabilized visual patterns still lack explicit clinical meaning. We therefore construct the visual-semantic representation in three stages: multi-granularity representation space, codebook-enhanced Hierarchical Visual Encoding and prompt-driven multi-granularity semantic generation.

\noindent\textbf{Multi-Granularity Representation Space.}
To reflect the hierarchical nature of ASD evidence, both modalities are represented at multiple granularities. For the visual modality, the encoder $\mathcal{E}_{vis}$ produces:
\begin{equation}
    \mathcal{V}=\mathcal{E}_{vis}(x)=\{\mathbf{v}^{(l)} \in \mathbb{R}^{H_l \times W_l \times D} \mid l=1,\dots,L\},
\end{equation}
where $\mathbf{v}^{(l)}$ denotes the visual representation at scale $l$. 

For the semantic modality, the generator $\mathcal{G}$ with hierarchical prompts $\mathcal{P}$ yields:
\begin{equation}
    \mathcal{T}=\mathcal{G}(x,\mathcal{P})=\{T_{k} \mid k \in \{\text{word},\text{phrase},\text{sent}\}\}.
\end{equation}
Here, $T_k$ denotes the generated diagnostic description at semantic granularity $k$. This formulation allows UniAR to associate visual evidence with semantic concepts at matched granularities.

\noindent \textbf{Codebook-Enhanced Hierarchical Visual Encoding.}
Directly aligning raw visual features with language is unreliable due to identity-specific appearance, background interference, and acquisition variation. We therefore first transform raw observations into more stable visual prototypes.

Given a visual observation, a multi-hierarchical encoder extracts feature maps at multiple scales: $\mathcal{V} = \{v_l\}_{l=1}^{L}, \quad v_l \in \mathbb{R}^{H_l \times W_l \times D}$,
where $v_l$ denotes the feature representation at scale $l$. To suppress redundant variations, we introduce a learnable \textit{Visual Codebook} $\mathcal{C} = \{\mathbf{c}_k\}_{k=1}^{K}$, where each codeword $\mathbf{c}_k \in \mathbb{R}^{D}$ represents an ASD-relevant visual prototype.

For each spatial feature vector $v_l^{(i,j)}$, nearest-neighbor quantization is performed as:
\begin{equation}
    z_q(v_l^{(i,j)}) = \mathbf{c}_{k}, \quad \text{where} \quad
    k = \arg\min_{n \in \{1,\dots,K\}} \|v_l^{(i,j)} - \mathbf{c}_n\|_2.
\end{equation}
$z_q(\cdot)$ denotes the quantization operator. By mapping continuous features into a discrete prototype space, the model suppresses instance-specific noise and encourages visually diverse samples to share ASD-salient behavioral patterns. Gradients are propagated through the quantization operation using the straight-through estimator.

\noindent \textbf{Prompt-Driven Multi-Granularity Semantic Generation.}
Although the visual codebook stabilizes the visual stream, the resulting prototypes remain diagnostically under-specified. To provide interpretable diagnostic anchors, we further construct hierarchical semantics from the visual input.

Because real ASD datasets rarely contain dense reports, we employ GPT-4o to generate descriptions at three complementary granularities: $\mathcal{T} = \{T_{\text{word}}, T_{\text{phrase}}, T_{\text{sent}}\}$, where $T_{\text{word}}$, $T_{\text{phrase}}$, and $T_{\text{sent}}$ denote word-level, phrase-level, and sentence-level descriptions, respectively. These levels progressively capture isolated attributes, local behavioral interactions, and holistic diagnostic context.

The generated texts are encoded by a frozen text encoder:
\begin{equation}
    \mathbf{S}_g = \mathrm{Enc}(T_g), \quad g \in \{\text{word}, \text{phrase}, \text{sent}\},
\end{equation}
yielding semantic embeddings $\mathbf{S}_{\text{word}}$, $\mathbf{S}_{\text{phrase}}$, and $\mathbf{S}_{\text{sent}}$. This hierarchical construction provides richer semantic anchors than a single global description and supports subsequent fine-grained cross-modal alignment.
\begin{figure}[t]
    \centering
    \includegraphics[width=\linewidth]{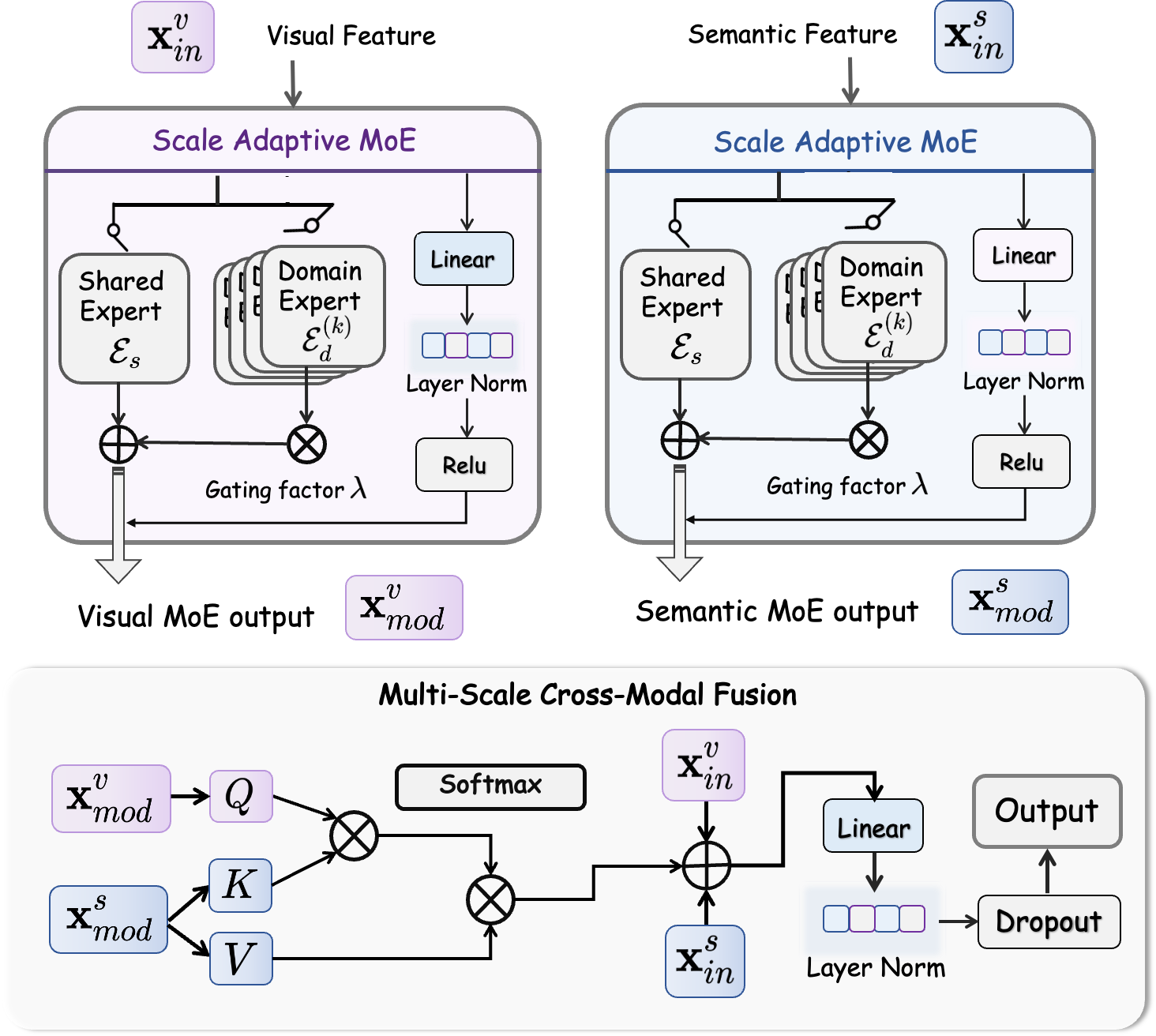}
    \caption{Detailed illustration of the Multi-Scale Alignment Module (MSAM) designed to tackle feature inconsistencies.}
    \Description{Three-stage multi-scale alignment module showing semantic and visual features aligned through cross-modal attention and fused for prediction.}
    \label{fig:msam_detail}
\end{figure}

\subsection{Multi-Scale Alignment Module}
\label{sec:alignment}
After prototype refinement and semantic generation, the model obtains more stable visual patterns and more informative diagnostic descriptions. However, these two modalities are still not directly compatible. In particular, visual prototypes and semantic embeddings differ not only in modality, but also in abstraction level: local visual cues should not interact with sentence-level semantics in the same manner as global behavioral patterns. As a result, naive concatenation or direct fusion may introduce severe cross-modal mismatch and weaken the benefit of hierarchical semantics. To address this issue, we propose a \textit{Multi-Scale Alignment Module (MSAM)} that explicitly performs scale-aware feature modulation and cross-modal interaction.

\noindent\textbf{Cross-Modal Alignment Objective.}
Since visual and semantic features come from heterogeneous spaces, UniAR learns a joint representation by minimizing their latent discrepancy:
\begin{equation}
    \min_{\theta}\sum_{l,k} D_{\phi}\Big(f_v(\mathbf{v}^{(l)}), f_t(\mathbf{t}^{(k)})\Big),
\end{equation}
where $f_v(\cdot)$ and $f_t(\cdot)$ are modality-specific projection functions, and $D_{\phi}$ measures their discrepancy. This motivates the scale-aware alignment design by encouraging matched visual and semantic representations to become more compatible in the latent space.

\noindent \textbf{Scale-Adaptive Modulation with MoE.}
Before multimodal fusion, the model should first reduce the feature discrepancy between different granularities. To this end, we introduce a scale-adaptive modulator implemented with a Mixture-of-Experts (MoE) mechanism. The main idea is to decompose each input feature into shared and scale-specific components, so that common diagnostic structure can be preserved while scale-dependent characteristics are selectively emphasized.

Let $\mathbf{x}_{in}$ denote the input feature for a specific alignment branch, which can be either a visual representation $\mathbf{x}_{in}^{v}$ or a semantic representation $\mathbf{x}_{in}^{s}$. We employ a \textit{Shared Expert} $\mathcal{E}_{s}(\cdot)$ to model common patterns across different scales, and a \textit{Scale-Specific Expert} $\mathcal{E}_{d}^{(k)}(\cdot)$ for the $k$-th granularity. The modulated feature is computed as:
\begin{equation}
    \mathbf{x}_{mod} = \mathcal{E}_{s}(\mathbf{x}_{in}) + \lambda \cdot \mathcal{E}_{d}^{(k)}(\mathbf{x}_{in}),
\end{equation}
where $\lambda$ is a learnable gating factor. Both experts are implemented as MLPs. This design enables the model to preserve cross-scale commonality while adaptively highlighting the semantic attributes that are most relevant to each granularity, thereby producing more compatible representations for subsequent alignment.

\begin{table*}[t]
    \centering
    \caption{Comparative experiments on MRI scenarios.} 
    \setlength{\tabcolsep}{3.5mm} 
    \renewcommand{\arraystretch}{0.8} 
    \definecolor{rowgray}{gray}{0.92} 
    \begin{small} 
    \begin{tabular}{lcccccccccccc}
    \toprule
    \multirow{2}{*}{\textbf{Method}} & \multicolumn{4}{c}{\textbf{ABIDE}} & \multicolumn{4}{c}{\textbf{ABIDE II}} & \multicolumn{4}{c}{\textbf{AVG}} \\
    \cmidrule(lr){2-5} \cmidrule(lr){6-9} \cmidrule(lr){10-13} 
     & Acc & Recall & Pre & AUC & Acc & Recall & Pre & AUC & Acc & Recall & Pre & AUC \\ 
    CPM      & 61.4 & 58.3 & 53.3 & 64.7 & 60.3 & 61.0 & 58.5 & 62.0 & 60.9 & 59.7 & 55.9 & 63.4 \\
    Braingnn & 60.2 & 60.4 & 50.3 & 58.2 & 58.2 & 59.3 & 50.2 & 57.3 & 59.2 & 59.9 & 50.3 & 57.8 \\
    SpectBGNN& 59.6 & 67.3 & 51.9 & 61.4 & 69.6 & \textbf{71.0} & 72.2 & 71.4 & 64.6 & 69.2 & 62.1 & 66.4 \\
    STAGIN   & 68.5 & 71.2 & 69.2 & 74.1 & 68.4 & 67.8 & 69.1 & 72.0 & 68.5 & 69.5 & 69.2 & 73.1 \\
    Bolt     & 67.2 & 68.4 & 66.3 & 71.3 & 70.2 & 70.5 & 71.4 & 75.0 & 68.7 & 69.5 & 68.9 & 73.2 \\
    Causality& 69.7 & 54.3 & 73.4 & 74.2 & 71.7 & 65.7 & \textbf{74.2} & 75.3 & 70.7 & 60.0 & 73.8 & 74.8 \\
    BrainNetMLP& 76.8 & 77.0 & 76.5 & 80.2 & 71.5 & 68.5 & 72.5 & 74.8 & 74.2 & 72.8 & 74.5 & 77.5 \\
    GPT-4o   & \underline{77.5} & \underline{78.2} & \underline{77.6} & \underline{81.3} & \underline{71.7} & 68.7 & 72.7 & 74.6 & \underline{74.4} & \underline{73.5} & \underline{75.2} & \underline{78.0} \\
    \rowcolor{rowgray} 
    \textbf{Ours} & \textbf{78.5} & \textbf{79.2} & \textbf{78.8} & \textbf{82.5} & \textbf{73.3} & \underline{69.4} & \underline{73.7} & \textbf{76.7} & \textbf{75.9} & \textbf{74.3} & \textbf{76.3} & \textbf{79.6} \\
    \bottomrule
    \end{tabular}
    \end{small}
    \label{tab:abide_all_comparison}
\end{table*}

\begin{table*}[t]
    \centering
    \caption{Comparative experiments on facial expression scenarios.}
    \setlength{\tabcolsep}{3.5mm} 
    \renewcommand{\arraystretch}{0.8} 
    \definecolor{rowgray}{gray}{0.92} 
    \begin{small} 
    \begin{tabular}{lcccccccccccc}
    \toprule
    \multirow{2}{*}{\textbf{Method}} & \multicolumn{4}{c}{\textbf{HRM}} & \multicolumn{4}{c}{\textbf{Kaggle}} & \multicolumn{4}{c}{\textbf{AVG}} \\
    \cmidrule(lr){2-5} \cmidrule(lr){6-9} \cmidrule(lr){10-13} 
     & Acc & Recall & Pre & AUC & Acc & Recall & Pre & AUC & Acc & Recall & Pre & AUC \\ 
    RAN  & 78.4 & 73.5 & 75.8 & 81.0 & \underline{98.2} & \underline{97.8} & 97.5 & \textbf{99.1} & 88.3 & 85.7 & 86.7 & 90.1 \\
    SCN  & 79.3 & 74.8 & 76.5 & 82.5 & 94.5 & 93.2 & 94.8 & 95.5 & 86.9 & 84.0 & 85.7 & 89.0 \\
    DMUE & 80.5 & 77.1 & 77.5 & 83.9 & 97.1 & 96.5 & 97.8 & 98.2 & 88.8 & 86.8 & 87.7 & 91.1 \\
    RUL  & 80.8 & 76.9 & 78.4 & 84.2 & 97.6 & 96.8 & \underline{98.1} & 98.5 & 89.2 & 86.9 & \underline{88.3} & 91.4 \\
    EAC  & 81.2 & 77.5 & 78.9 & 84.8 & 96.3 & 95.9 & 96.7 & 97.5 & 88.8 & 86.7 & 87.8 & 91.2 \\
    GPT-4o & \underline{82.6} & \underline{79.2} & \underline{80.3} & \underline{85.7} & 98.2 & 97.2 & 97.8 & \underline{98.7} & \underline{90.4} & \underline{88.2} & 89.1 & \underline{92.1} \\
    \rowcolor{rowgray} 
    \textbf{Ours} & \textbf{84.5} & \textbf{81.2} & \textbf{81.8} & \textbf{87.6} & \textbf{98.7} & \textbf{97.9} & \textbf{98.2} & 98.7 & \textbf{91.6} & \textbf{89.6} & \textbf{90.0} & \textbf{93.2} \\
    \bottomrule
    \end{tabular}
    \end{small}
    \label{tab:hrm_kaggle_comparison}
\end{table*}
\noindent \textbf{Multi-Scale Cross-Modal Fusion.}
Once the features are modulated into a more compatible latent space, the next challenge is to establish effective interactions between matched visual and semantic granularities. We therefore adopt a multi-scale cross-attention mechanism to inject diagnostic semantics into visual prototypes in a scale-aware manner.

For each branch, the modulated visual features are treated as Queries, while the corresponding semantic embeddings serve as Keys and Values. The cross-modal interaction is defined as:
\begin{equation}
    \text{Attention}(Q,K,V) =
    \text{Softmax}\left(\frac{\mathbf{x}_{mod}^{v}(\mathbf{x}_{mod}^{s})^{T}}{\sqrt{d_k}}\right)\mathbf{x}_{mod}^{s},
\end{equation}
where $\mathbf{x}_{mod}^{v}$ and $\mathbf{x}_{mod}^{s}$ denote the modulated visual and semantic features, respectively. Based on this interaction, the aligned representation is obtained by:
\begin{equation}
    \mathbf{Z} = \text{LN}\left(\mathbf{x}_{in}^{v} + \mathbf{x}_{in}^{s} + \text{Attention}(Q,K,V)\right),
\end{equation}
where $\text{LN}(\cdot)$ denotes Layer Normalization.

This process progressively injects hierarchical diagnostic semantics into the visual stream, allowing the model to connect local visual biomarkers with fine-grained linguistic cues and global visual patterns with holistic diagnostic context. Finally, the aligned outputs from all branches are aggregated to form the final representation for ASD recognition.

\section{Experiments}
\subsection{Experimental Setting}
\label{subsec:exp_setup}
\noindent\textbf{Public Benchmarks.}
Autism-related biomarkers in our study are manifested in two complementary modalities, \textit{i.e.}, brain MRI connectivity and facial behavioral expressions. To comprehensively evaluate the robustness and generalization of UniAR, we conduct experiments on four public benchmarks together with one self-constructed cross-platform social-media benchmark. \textit{MRI Scenario (ABIDE I \& II):} We evaluate on the Autism Brain Imaging Data Exchange benchmarks. ABIDE I contains 1,112 subjects (539 ASD and 573 controls) from 17 sites, serving as a standard multi-site benchmark for ASD identification. ABIDE II includes 1,114 subjects from 19 sites and provides a more challenging setting with greater inter-site variability. \textit{Facial Expression Scenario (HRM \& Kaggle):} We use two facial benchmarks to cover both temporal and static behavioral cues. The HRM dataset contains 131,758 frames from 1,535 videos for evaluating subtle temporal facial dynamics. In addition, we use an aggregated static facial benchmark collected from Kaggle and Zenodo, containing 12,381 images, to assess generalization.

\begin{figure}[t]
    \centering
    \includegraphics[width=1.0\linewidth]{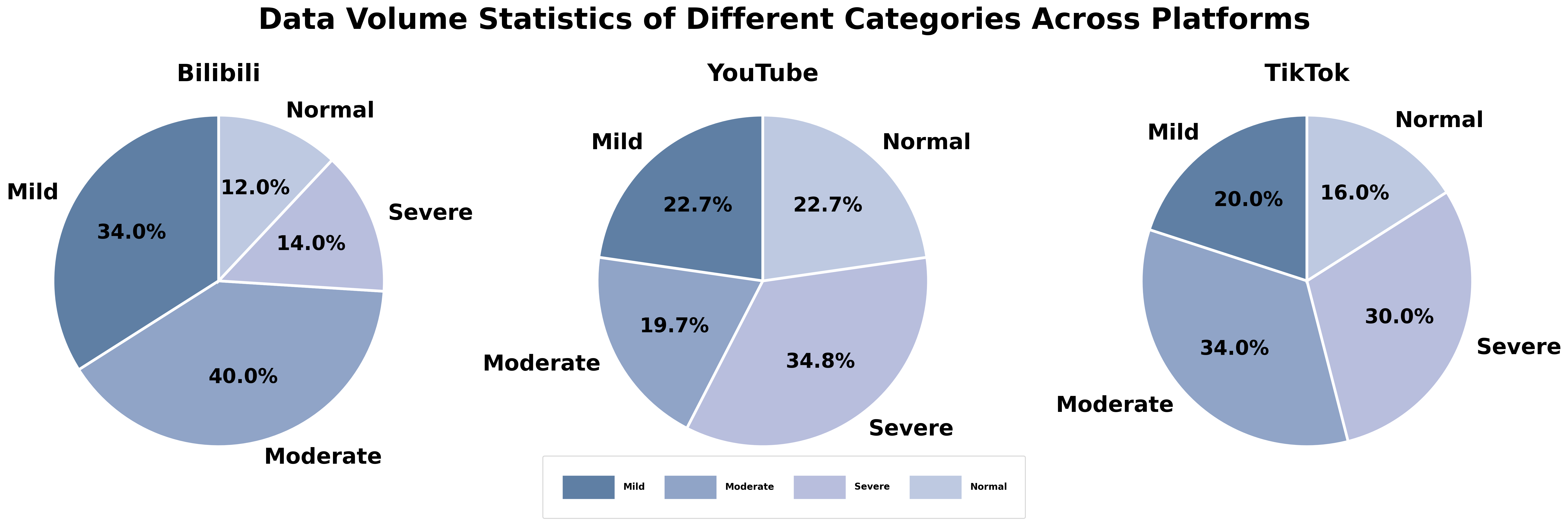}
    \caption{Data volume statistics of different categories across platforms.}
    \Description{Pie charts showing the distribution of normal, mild, moderate, and severe samples across Bilibili, YouTube, and TikTok.}
\label{fig:platform_category_pie_chart}
\end{figure}

\noindent\textbf{ASD-MM Dataset Construction and Evaluation Protocol.}To complement the closed binary ASD-vs-control settings of existing public benchmarks, we further construct the first cross-platform social-media autism benchmark, termed ASD-MM, for finer-grained autism assessment under realistic distribution shifts. ASD-MM consists of autism-related videos paired with expert-associated diagnostic descriptions and defines autism-related assessment as a four-class problem with labels \textit{Normal}, \textit{Mild}, \textit{Moderate}, and \textit{Severe}. The videos are collected from three mainstream social-media platforms, namely Bilibili, YouTube, and TikTok. After data cleaning and screening, representative frames are manually extracted from each video for evaluation. All samples are annotated by three clinical practitioners, and each sample is associated with an expert-written diagnostic description covering three dimensions: social competence, language expression, and behavior. Category statistics of ASD-MM are summarized in Figure~\ref{fig:platform_category_pie_chart}. In experiments, ASD-MM serves as an external validation benchmark for both cross-platform four-class classification and quantitative interpretability evaluation. 

\noindent\textbf{Baselines.}
We compare UniAR against representative methods designed for the two visual scenarios.\textit{For MRI scenarios}, we include methods from six paradigms:
(i) \textit{graph-based learning}, including BrainGNN~\cite{li2021braingnn}, STAGIN~\cite{kim2021learning}, and Spectral BGNN~\cite{duan2024spectral};
(ii) \textit{temporal modeling}, represented by BolT~\cite{bedel2023bolt};
(iii) \textit{classical statistical prediction}, represented by CPM~\cite{shen2017using};(iv) \textit{causal modeling}, represented by the causality-inspired LSTM framework~\cite{duan2025causal};and(v) \textit{MLP-based modeling}, represented by BrainNetMLP~\cite{hou2025brainnetmlp},an efficient baseline for functional brain network classification.
\textit{For facial expression scenarios}, we compare with robust CNN-based, vision-language, and instruction-tuned methods, including SCN~\cite{wang2020suppressing}, DMUE~\cite{she2021dive}, RUL~\cite{zhang2021relative}, EAC~\cite{zhang2022learn}, RAN~\cite{wang2020region}.
For the cross-platform ASD-MM benchmark, we compare with EAC under the same eval protocol.

\noindent\textbf{Metrics.}
For the public classification benchmarks, we report Accuracy, Precision, Recall, and Area Under the ROC Curve (AUC), which measure diagnostic correctness and discriminative ability. For the ASD-MM benchmark, we report Accuracy, F1-score, and Recall for four-class classification performance. To evaluate the quality of generated diagnostic descriptions, we further adopt ROUGE-1, ROUGE-2, and ROUGE-L against expert-written reports.

\noindent\textbf{Implementation Details.}
Our framework is implemented in PyTorch and trained on a single NVIDIA A100 GPU. For ABIDE I and II, we extract functional connectivity patterns as model input, while the facial datasets undergo standard visual preprocessing. Multi-Granularity prompts are generated using GPT-4o and then encoded by a frozen BERT encoder. For visual discretization, we use a multi-scale vector quantization module with a codebook size of 128 and an embedding dimension of 256. The quantized representations are projected back to the original feature space through a linear layer for dimensional consistency. In the MoE branch, each routing branch contains 4 experts, with hidden dimension 512 and input/output dimension 256. The gating network uses a 64-dimensional hidden layer to generate softmax routing coefficients. We train the model using the Adam optimizer with a learning rate of $1\times10^{-4}$ and a batch size of 32 for 30 epochs, with early stopping. For the public benchmarks, the train/test ratio is set to 7:3.

\subsection{Comparison Results}
\label{subsec:comparison_results}

\noindent\textbf{Performance on MRI Scenarios.}
Table~\ref{tab:abide_all_comparison} reports the results on the heterogeneous ABIDE benchmarks. UniAR achieves the best overall performance on both ABIDE and ABIDE II, ranking first on all four metrics for ABIDE and obtaining the best Accuracy and AUC on ABIDE II. Compared with ASD-specific baselines, including CPM, Braingnn, SpectBGNN , STAGIN, Bolt and Causality, the advantage of UniAR is clear. In particular, relative to the causality-based model, UniAR improves Recall from 54.3\% to 79.2\% on ABIDE and from 60.0\% to 74.3\% on the average results, showing a much better balance between sensitivity and precision for screening-oriented ASD recognition. We further include BrainNetMLP and GPT-4o as stronger generic and foundation-model baselines, and UniAR still remains superior in overall Accuracy and AUC, indicating that the gain does not come merely from generated semantics, but from the joint effect of the multi-granularity visual-semantic representation construction module and the multi-scale alignment module.

\noindent\textbf{Performance on Facial Expression Scenarios.}
Table~\ref{tab:hrm_kaggle_comparison} reports the results on the facial expression benchmarks. UniAR consistently achieves the best overall performance across both dynamic and static settings. Compared with ASD/facial-specific baselines, including RAN, SCN, DMUE, RUL, and EAC, the advantage of UniAR is clear on both HRM and Kaggle. In particular, on the dynamic HRM benchmark, UniAR reaches 84.5\% Accuracy, outperforming EAC by 3.3 percentage points and GPT-4o by 1.9 percentage points, while also ranking first on Recall, Precision, and AUC. On the average results, UniAR improves Accuracy from 88.8\% to 91.6\% over EAC and from 90.4\% to 91.6\% over GPT-4o. These results suggest that the gain of UniAR does not come solely from generated descriptions, but from the joint effect of the multi-granularity visual-semantic representation construction module and the multi-scale alignment module. The consistent gains across HRM and Kaggle further indicate that the proposed framework is effective for ASD-related facial behavior understanding across both dynamic and static scenarios.

\begin{table*}[t]
    \centering
    \caption{Ablation study on MRI scenarios.}
    \setlength{\tabcolsep}{1.6mm}
    \renewcommand{\arraystretch}{0.9}
    \definecolor{rowgray}{gray}{0.92}
    
    \resizebox{1\linewidth}{!}{
    \begin{tabular}{lcccccccccccccccc}
    \toprule
    \multirow{2}{*}{\textbf{Method}} & \multicolumn{4}{c}{\textbf{Modules}} & \multicolumn{4}{c}{\textbf{ABIDE}} & \multicolumn{4}{c}{\textbf{ABIDE II}} & \multicolumn{4}{c}{\textbf{AVG}} \\
    \cmidrule(lr){2-5} \cmidrule(lr){6-9} \cmidrule(lr){10-13} \cmidrule(lr){14-17}
     & Word & Phrase & Sent. & Fusion & Acc & Recall & Pre & AUC & Acc & Recall & Pre & AUC & Acc & Recall & Pre & AUC \\
    \midrule
    Baseline & $\times$ & $\times$ & $\times$ & $\times$ & 76.2 & 77.1 & 76.5 & 80.1 & 69.1 & 65.8 & 70.2 & 72.3 & 72.7 & 71.5 & 73.4 & 76.2 \\
    w/ Word & $\checkmark$ & $\times$ & $\times$ & $\times$ & 76.9 & 77.6 & 77.2 & 80.7 & 70.0 & 66.9 & 71.5 & 73.1 & 73.5 & 72.3 & 74.4 & 76.9 \\
    w/ Word, Phrase & $\checkmark$ & $\checkmark$ & $\times$ & $\times$ & 77.6 & 78.3 & 78.0 & 81.4 & 71.0 & 68.1 & 72.6 & 74.2 & 74.3 & 73.2 & 75.3 & 77.8 \\
    w/ Word, Phrase, Sentence & $\checkmark$ & $\checkmark$ & $\checkmark$ & $\times$ & 78.1 & 78.8 & 78.5 & 82.0 & 71.8 & 68.8 & 73.2 & 75.0 & 75.0 & 73.8 & 75.9 & 78.5 \\
    \rowcolor{rowgray} 
    \textbf{Ours} & \textbf{$\checkmark$} & \textbf{$\checkmark$} & \textbf{$\checkmark$} & \textbf{$\checkmark$} & \textbf{78.5} & \textbf{79.2} & \textbf{78.8} & \textbf{82.5} & \textbf{73.3} & \textbf{69.4} & \textbf{73.7} & \textbf{76.7} & \textbf{75.9} & \textbf{74.3} & \textbf{76.3} & \textbf{79.6} \\
    \bottomrule
    \end{tabular}
    }
    \label{tab:mri_ablation}
\end{table*}

\begin{table*}[t]
    \centering
    \caption{Experimental results of different codebook sizes on multiple datasets.}
    \label{tab:codebook_size_final}
    \setlength{\tabcolsep}{1.2mm}  
    \footnotesize  
    \resizebox{\linewidth}{!}{\begin{tabular}{l|cccc|cccc|cccc|cccc}
    \toprule
    \multirow{2}{*}{\textbf{Codebook Size}} & \multicolumn{4}{c|}{\textbf{HRM}} & \multicolumn{4}{c|}{\textbf{Kaggle}} & \multicolumn{4}{c|}{\textbf{ABIDE}} & \multicolumn{4}{c}{\textbf{ABIDE II}} \\
    \cline{2-17}
    & Accuracy & Recall & Precision & AUC & Accuracy & Recall & Precision & AUC & Accuracy & Recall & Precision & AUC & Accuracy & Recall & Precision & AUC \\ 
    \midrule  
    64 & 83.8 & 80.5 & 81.1 & 86.9 & 98.4 & 97.6 & 97.9 & 98.1 & 78.1 & 78.6 & 78.2 & 81.9 & 71.8 & 68.9 & 73.2 & 75.2 \\
    \rowcolor{rowgray}128 & \textbf{84.5} & \textbf{81.2} & \textbf{81.8} & \textbf{87.6} & \textbf{98.7} & \textbf{97.9} & \textbf{98.2} & \textbf{98.7} & \textbf{78.5} & \textbf{78.9} & \textbf{78.8} & \textbf{82.5} & \textbf{72.3} & \textbf{69.4} & \textbf{73.7} & \textbf{75.7} \\
    256 & 84.2 & 81.4 & 81.5 & 87.5 & 98.5 & 97.7 & 98.1 & 98.8 & 78.3 & 78.9 & 78.6 & 82.6 & 72.1 & 69.3 & 73.5 & 75.8 \\
    \bottomrule
    \end{tabular}}
    \vspace{0.5em}
\end{table*}

\begin{table*}[t]
    \centering
    \caption{Comparative experiments with different numbers of experts on multiple datasets.}
    \label{tab:expert_num_comparison}

    \footnotesize  
    \resizebox{\textwidth}{!}{%
    \begin{tabular}{l|cccc|cccc|cccc|cccc}
    \toprule
    \multirow{2}{*}{\textbf{Number of Experts}} & \multicolumn{4}{c|}{\textbf{HRM}} & \multicolumn{4}{c|}{\textbf{Kaggle}} & \multicolumn{4}{c|}{\textbf{ABIDE}} & \multicolumn{4}{c}{\textbf{ABIDE II}} \\
    \cline{2-17}
    & Accuracy & Recall & Precision & AUC & Accuracy & Recall & Precision & AUC & Accuracy & Recall & Precision & AUC & Accuracy & Recall & Precision & AUC \\ 
    \midrule  
    2 & 84.2 & 80.9 & 81.5 & 87.2 & 98.4 & 97.7 & 97.9 & 98.4 & 78.2 & 78.8 & 78.6 & 82.2 & 72.2 & 69.2 & 73.5 & 75.3 \\
    3 & 84.4 & 81.2 & 81.7 & 87.4 & 98.7 & 97.7 & 98.1 & 98.6 & 78.4 & 78.7 & 78.9 & 82.3 & 72.1 & 69.2 & 73.6 & 75.5 \\
    \rowcolor{rowgray}4 & \textbf{84.5} & \textbf{81.2} & \textbf{81.8} & \textbf{87.6} & \textbf{98.7} & \textbf{97.9} & \textbf{98.2} & \textbf{98.7} & \textbf{78.5} & \textbf{79.2} & \textbf{78.8} & \textbf{82.5} & \textbf{72.3} & \textbf{69.4} & \textbf{73.7} & \textbf{75.7} \\
    \bottomrule
    \end{tabular}
    }
    \vspace{0.5em}  
\end{table*}

\subsection{Ablation Study}
\label{subsec:ablation_study}

\begin{figure}[t]
    \centering
    \includegraphics[width=1\linewidth]{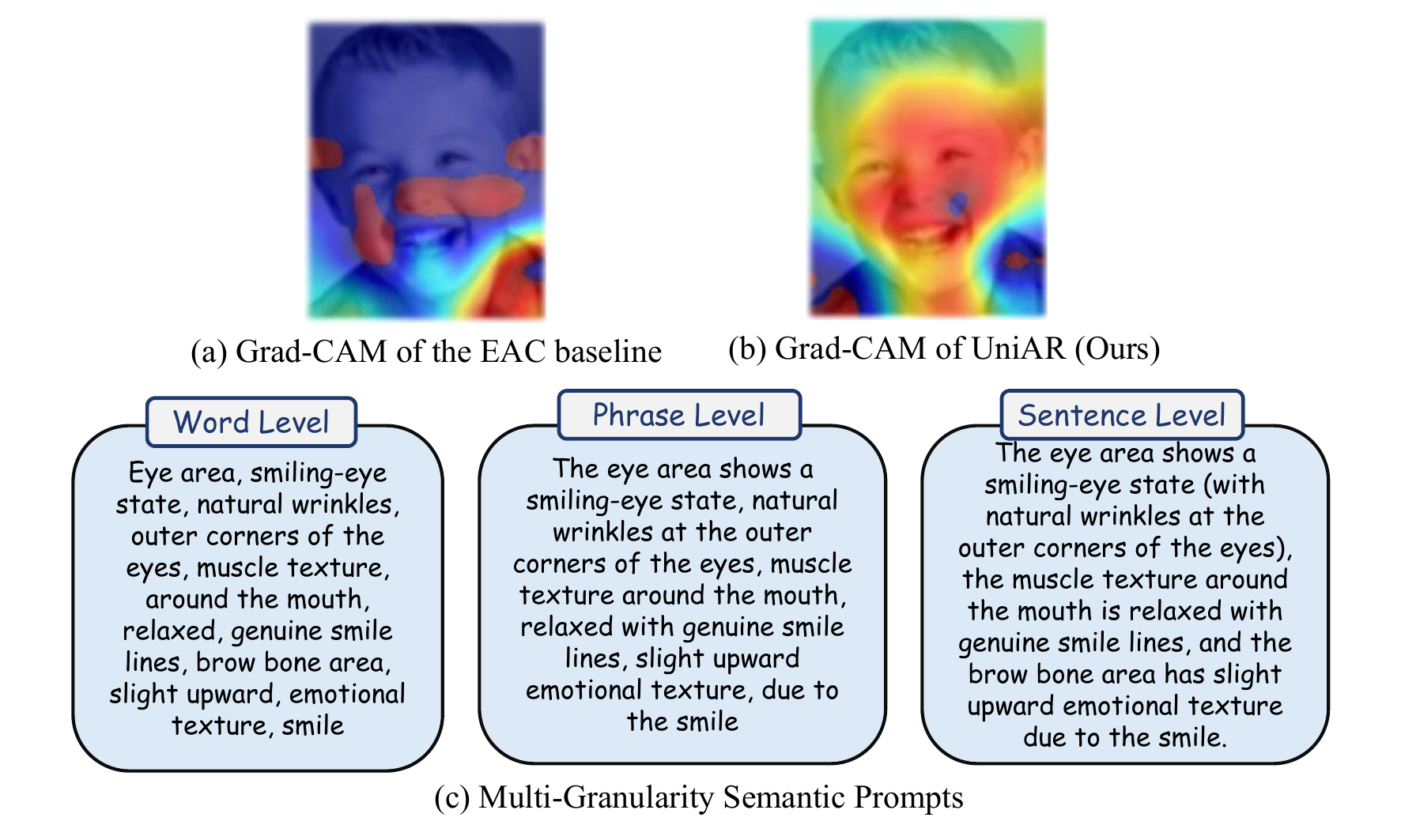}
    \vspace{-10pt}
    \caption{Qualitative visualization of attention mechanisms and hierarchical semantic alignment.}
    \label{fig:qualitative_vis}
\end{figure}
We conduct ablation and design analysis from two perspectives. First, we examine the effectiveness of the proposed hierarchical alignment and fusion mechanism by progressively adding the word-level, phrase-level, sentence-level, and fusion modules. Second, we study whether two key architectural choices---the codebook size and the number of experts in the MoE branch---are reasonably selected. These aim to verify not only that the proposed components are effective, but also that the overall design of UniAR is well aligned with the demands of ASD recognition under semantic scarcity.

\noindent\textbf{Module Ablation on MRI Scenarios.}
We first evaluate the contribution of each semantic alignment component and the fusion module on the MRI benchmarks, as shown in Table~\ref{tab:mri_ablation}. Starting from the baseline (\textit{w/o All}), performance improves steadily as word-level, phrase-level, and sentence-level alignment are introduced in sequence. This trend suggests that hierarchical semantic guidance progressively enriches the model's understanding of ASD-related patterns, moving from fine-grained local semantics to more holistic diagnostic context. In particular, the addition of sentence-level alignment brings a clear improvement over the word+phrase setting, indicating that global semantic context is helpful for interpreting heterogeneous MRI biomarkers. After further adding the fusion module, the full model achieves the best overall performance, which supports the role of fusion in integrating multi-scale visual-semantic evidence. Similar observations can also be found on the facial expression benchmarks. 

\noindent\textbf{Sensitivity to Codebook Size.}
We next investigate the influence of codebook size on UniAR, as summarized in Table~\ref{tab:codebook_size_final}. Since the visual codebook is responsible for quantizing visual features into more stable prototypes, its size directly affects the balance between representation diversity and prototype compactness. The results show that a codebook size of 128 achieves the best overall performance across the four datasets. A smaller codebook (64) appears insufficient to capture diverse ASD-related visual patterns, while a larger one (256) introduces limited or inconsistent gains, suggesting that excessive prototype capacity may weaken representation compactness and stability. These results support the choice of a moderate codebook size for robust multi-scale visual abstraction.

\noindent\textbf{Sensitivity to the Number of Experts.}
We further analyze the effect of the number of experts in the MoE-based modulation branch in Table~\ref{tab:expert_num_comparison}, which shows that increasing the number of experts from 2 to 4 consistently improves performance across most datasets, and the best overall results are achieved with 4 experts. This suggests that a moderate increase in expert diversity is beneficial for modeling heterogeneous visual-semantic patterns, especially when aligning features from different abstraction levels. In contrast, smaller expert configurations provide limited routing flexibility and are less effective at handling the complexity of cross-dataset variations.

\subsection{Qualitative Analysis and Visualization}
\label{subsec:qualitative}
To qualitatively examine the interpretability of UniAR, we use Grad-CAM to project decision activations back to the input space, where warmer colors indicate higher relevance. As shown in Figure~\ref{fig:qualitative_vis}, UniAR exhibits two visible advantages over the EAC baseline. First, its high-response regions are more concentrated on clinically relevant facial areas, such as the periorbital and perioral regions, with substantially less attention leakage to irrelevant background regions. Second, the activation patterns of UniAR appear more spatially coherent, whereas the baseline shows more scattered and fragmented responses. These observations suggest that UniAR establishes a more reliable correspondence between facial evidence and diagnostically relevant semantics.

These qualitative patterns are also consistent with the design of our hierarchical visual-semantic alignment. Word-level prompts provide fine-grained semantic anchors for local facial cues, phrase-level prompts capture relations among facial components and behavioral patterns, and sentence-level prompts introduce more global diagnostic context. Although Grad-CAM does not provide direct causal evidence for each semantic level, the improved localization and coherence of UniAR support the effectiveness of aligning visual evidence with multi-granularity diagnostic semantics.

\subsection{Cross-Platform Quantitative Validation and Interpretable Analysis}
To further evaluate the practical generalization of UniAR beyond closed public benchmarks, we conduct additional experiments on the self-constructed ASD-MM benchmark. In particular, this setting is designed to assess two aspects: whether UniAR remains robust under cross-platform distribution shifts, and whether the generated diagnostic descriptions remain semantically consistent with expert-associated annotations in realistic scenarios. 

\begin{figure}[t]
    \centering
    \includegraphics[width=1.0\linewidth]{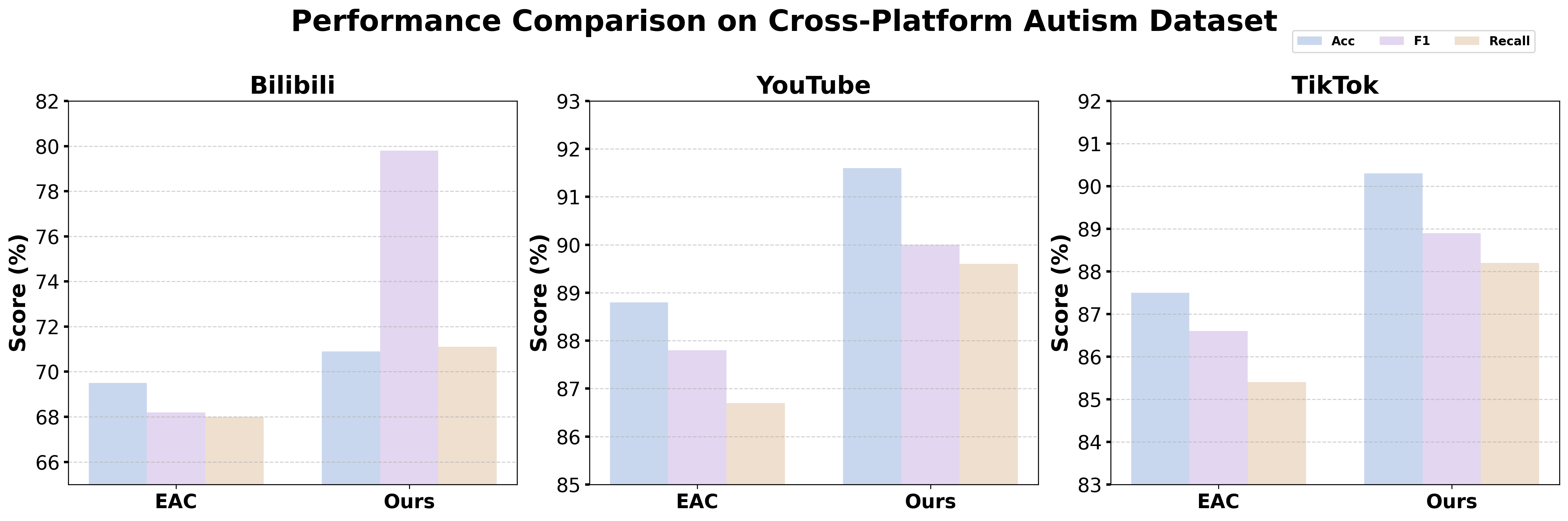}
    \caption{Cross-platform performance comparison on social media datasets.}
    \Description{Bar charts comparing UniAR and the EAC baseline on accuracy, F1-score, and recall across three social-media platforms.}
    \label{fig:cross_platform_performance}
\end{figure}

\noindent\textbf{Cross-Platform Four-Class Classification.}
We first evaluate UniAR on ASD-MM under the four-class setting of \textit{Normal}, \textit{Mild}, \textit{Moderate}, and \textit{Severe}. As shown in Figure~\ref{fig:cross_platform_performance}, UniAR consistently outperforms the EAC baseline across the evaluated social-media platforms in terms of Accuracy, F1-score, and Recall. These results indicate that the proposed visual-semantic alignment strategy generalizes more robustly than a strong task-specific baseline under realistic cross-platform distribution shifts.

\noindent\textbf{Quantitative Interpretability Evaluation.}
Since the diagnostic descriptions used in UniAR are generated rather than directly observed, it is important to examine whether they remain semantically faithful to expert knowledge instead of introducing unreliable hallucinated content. To this end, we further compare the generated descriptions of UniAR with expert-associated diagnostic reports using ROUGE-1, ROUGE-2, and ROUGE-L. As reported in Table~\ref{tab:cross_platform_metrics}, UniAR achieves favorable semantic consistency with expert reports across all evaluated platforms. These results suggest that the generated diagnostic descriptions capture diagnostically meaningful content to a reasonable extent, thereby providing additional quantitative support for the interpretability of UniAR in realistic deployment settings. 

\begin{table}[t]
    \centering
    \caption{Evaluation metrics on cross-platform datasets.}
    \small
    \setlength{\tabcolsep}{4pt}
    \begin{tabular}{l|ccc}
        \toprule
        Platform & Rouge-1 & Rouge-2 & Rouge-L \\ 
        \hline
        Bilibili & 72.3 & 65.7 & 70.1 \\
        YouTube  & 78.6 & 72.4 & 76.9 \\
        TikTok   & 77.8 & 71.3 & 76.2 \\
        \bottomrule
    \end{tabular}
    \label{tab:cross_platform_metrics}
\end{table}

\section{Conclusion}
This paper addresses the challenge of semantic scarcity in ASD recognition by proposing UniAR, a unified framework that reconstructs hierarchical diagnostic semantics through multi-granularity prompt learning and aligns them with visual evidence in a scale-aware manner. Extensive experiments show that structured semantic supervision is highly beneficial for robust ASD recognition, especially under heterogeneous MRI settings, subtle facial behavior variations, and incomplete diagnostic text conditions. By combining prototype-based visual representation refinement with hierarchical visual-semantic alignment, UniAR effectively improves both robustness and interpretability, providing a practical way to bridge low-level biomarkers and high-level diagnostic reasoning.



\newpage
\bibliographystyle{ACM-Reference-Format}
\bibliography{references}


\end{document}